\documentclass[conference]{IEEEtran}
\IEEEoverridecommandlockouts
\usepackage{cite}
\usepackage{amsmath,amssymb,amsfonts}
\usepackage{algorithmic}
\usepackage{graphicx}
\usepackage{textcomp}
\usepackage{xcolor}
\usepackage{url}
\usepackage[hidelinks]{hyperref}
\usepackage{placeins}
\usepackage{float}
\usepackage{lineno}
\usepackage[english]{babel}
\def\BibTeX{{\rm B\kern-.05em{\sc i\kern-.025em b}\kern-.08em
    T\kern-.1667em\lower.7ex\hbox{E}\kern-.125emX}}
\begin{document}

\title{Evaluating the printability of stl files with ML}

\author{
\IEEEauthorblockN{Janik Henn}
\IEEEauthorblockA{\textit{MIT-KIT} 
Germany 
}
\and
\IEEEauthorblockN{Adrian Hauptmannl}
\IEEEauthorblockA{\textit{MIT-KIT}
Germany 
}
\and
\IEEEauthorblockN{Hamza A. A. Gardi}
\IEEEauthorblockA{\textit{IIIT at ETIT- KIT}
Germany 
}

}

\maketitle

\begin{abstract}
3D printing has long been a technology for industry professionals and enthusiasts willing to tinker or even build their own machines. This stands in stark contrast to today’s market, where recent developments have prioritized ease of use to attract a broader audience. Slicing software nowadays has a few ways to sanity check the input file as well as the output gcode. Our approach introduces a novel layer of support by training an AI model to detect common issues in 3D models. The goal is to assist less experienced users by identifying features that are likely to cause print failures due to difficult to print geometries before printing even begins.
\end{abstract}


\section{Introduction} \label{Introduction}

Fused Deposition Modeling (FDM) 3D printing is a complex, layer-by-layer manufacturing process, typically involving layers around $0.2$mm thick. Due to this, there are several scenarios in which small deviations are enough to cause a failed part. For example, when a new layer is deposited, the underlying layer has already cooled significantly. As the newly deposited layer contracts during cooling, it may cause enough stress to bend the part upward. Such issues are often the result of geometries that are not well suited to the constraints of the FDM process.

Many of these problems can be mitigated if the part is designed with process limitations in mind or if the user is aware of potential issues during slicing, allowing for appropriate adjustment of print settings. However, this knowledge is usually gained through experience, which means that inexperienced users or users experienced with other manufacturing processes are more prone to failures.

To assist users, modern slicing software does some basic assessment of the 3D model. For instance, warnings are shown when the contact area with the build plate is too small, or even brims may be added automatically. This increases the surface area by printing additional lines around the part. While very effective, this approach requires to define clear thresholds for extracted geometric properties for each possible issue. 

In this work, we propose a more general approach: training an AI model to identify common failure risks in 3D models. Instead of manually defining conditions for each individual issue, our method uses learned patterns to detect problematic geometries and assess their severity. The goal is to create a workflow that analyzes a given STL mesh and outputs a ranked list of potential issues based on their significance.

To achieve this, the model architecture must be capable of interpreting 3D geometry. A common method involves converting the mesh into a voxel grid, which can be processed using convolutional neural networks (CNNs), similar to those used in image analysis. However, due to the high resolution required by our application, we instead opted for the PointNet architecture. In this approach, each mesh is sampled as a fixed number of points, meaning that the input part size does not affect the number of useful data points fed into the model.

\begin{figure}[ht!]
        \centerline{\includegraphics[width=0.5\textwidth]{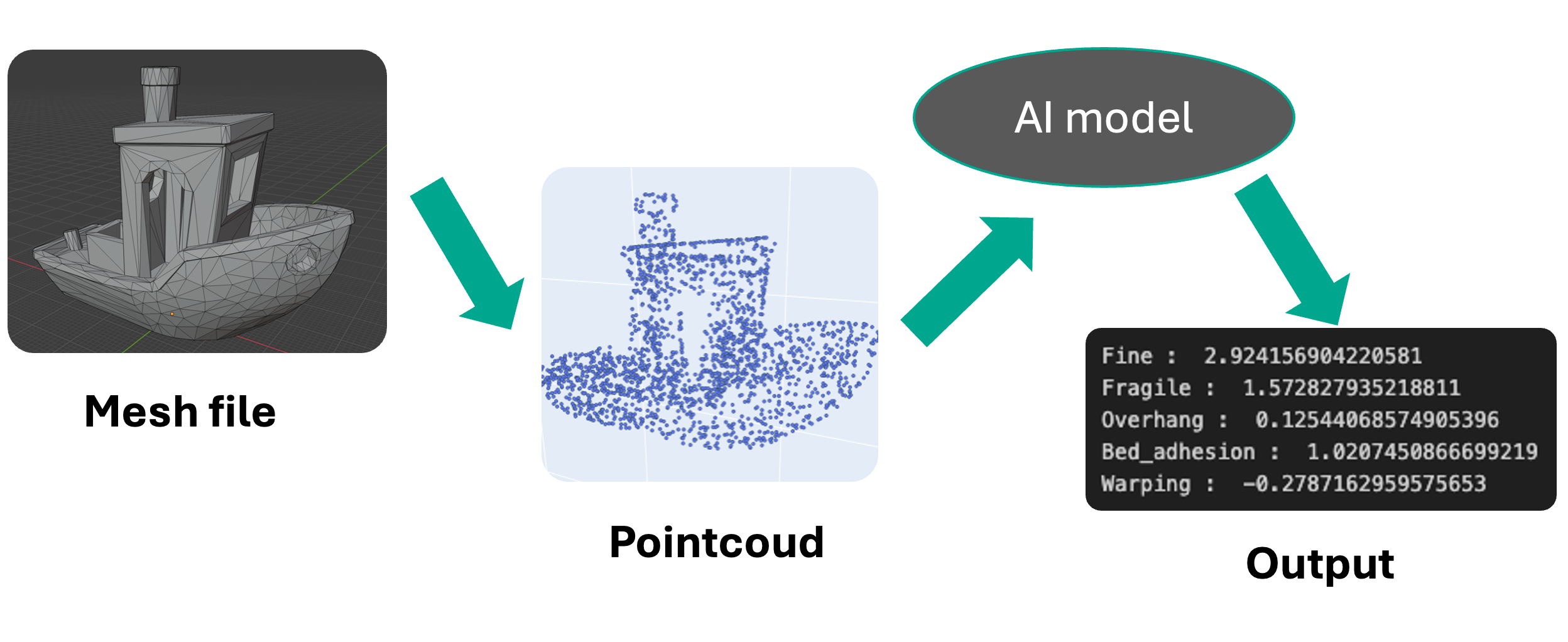}}
        \caption{Illustration of steps to our approach}
        \label{fig:approach}
    \end{figure}

Since our approach is relatively novel, we created a custom dataset for training and validation. The dataset consists of approximately 150 3D models, with around 20 reserved for validation and some of the remainder used for training. As none of the 3D models used for training were physically printed and tested, the labels may reflect subjective judgments based on our own experience, introducing potential bias. To mitigate this, model validation was conducted not only with the reserved validation subset but also through a small survey. In this survey, other 3D printing users were asked to independently assess and label a selection of models, allowing us to compare their evaluations with those of the AI.

\section{Related work}
    \subsection{Evaluating printability}
    
    As the 3D printing market rapidly expanded over the last decade, so has the research. A key topic is improving reliability and therefore driving operating costs down. This is why some research focuses on educating novice users in particular about the limitations of additive manufacturing and their implications on model design. Notably, a study found that users without prior training experienced a 33\% print failure rate, compared to just 6\% among those who had reviewed training materials \cite{10.1115/1.4037251}.

    One practical tool aimed at addressing this issue is the design checklist introduced by Booth et al \cite{10.1115/1.4037251}. This guideline points out geometries that are likely to cause printing problems, such as steep overhangs, thin features, and hard-to-remove support structures. Beyond geometry, the checklist encourages users to consider whether additive manufacturing is the appropriate method for producing a given part. Users rate the relevance of each category to their design, and the resulting scores are summed to provide an overall assessment. This score can then be compared to predefined thresholds to evaluate the printability risk of the part.

    A more analytical method is presented by Fudos et al. \cite{DBLP:journals/corr/abs-2010-12930}, who propose a probabilistic approach to assessing printability. Like Booth et al., they begin by identifying critical part characteristics that affect print success, such as structural robustness, the presence of fine details, support requirements, and overhangs. Their method evaluates printability using two probability functions. The first function considers general factors including part complexity, intended application, and the chosen 3D printing process. This allows for an informed comparison between manufacturing methods. The second function focuses on specific geometric features, weighting them based on size, application, and process compatibility. While this model incorporates a broad range of variables and offers a flexible framework, the paper does not provide detailed guidance on how to preprocess and segment 3D model data for use with the probability functions. This implies, that extracting the data for rating a parts printability is a very manual process and one that may introduce some subjectivity.   

    \subsection{Comparing approaches for 3D object classification}
    The field of 3D point cloud processing has seen significant advances in recent years, particularly through the application of deep learning techniques. As surveyed in \cite{ZHANG2023102456}, deep learning approaches for point cloud classification can be broadly categorized into multi-view, voxel-based and point-based methods.
    Multi-view approaches, such as MVCNN, project 3D models into multiple 2D images and leverage conventional 2D CNNs. While effective in some cases, these methods often suffer from information loss during projection and are not ideally suited for fine-grained geometric analysis. This is especially problematic because certain internal 3D-structures may not be picked up at all.  Voxel-based methods (e.g., VoxNet, OctNet) discretize the 3D space into a regular grid, enabling the use of 3D CNNs. However, they are computationally expensive and limited in resolution due to memory constraints. 
    Point-based architectures, beginning with PointNet \cite{charles_pointnet_2017}, directly operate on raw point clouds, preserving spatial granularity and permutation invariance. This class of models has since evolved to include more advanced methods like PointNet++, DGCNN, and transformer-based models (e.g., Point-BERT, Point-MAE), which even further increase accuracy, coming with drawbacks of high memory usage and general computational expense.
    Even further pushing the limits is the recent PointGST~\cite{Pointgst}. Being currently state of the art on Benchmarks like ModelNet40~\cite{ModelNet} PointGST is a transformer based model focusing on efficient fine tuning with performant results. 
    Despite this progress, most existing models are evaluated on general object classification or segmentation tasks and not specifically designed to assess 3D printability, this is a domain with distinct requirements such as robustness to small geometric features, sensitivity to overhangs, and further constraints. Our work aims to fill this gap by adapting point-based deep learning methods, particularly the PointNet architecture, for the novel task of evaluating 3D models with respect to their manufacturability via FDM 3D printing.   

\section{Brief introduction to FDM 3D-printing} \label{intro to 3d printing}

FDM stands for Fused Deposition Modeling, a 3D printing process in which an object is created by depositing molten material layer by layer. The material, typically a thermoplastic filament, is pulled into the extruder from a spool. The extruder pushes the filament through a heated element where it melts and then comes out of a fine nozzle. The nozzle traces each layer by moving relative to the print bed while continuously extruding material.

\begin{figure}[ht!]
        \centerline{\includegraphics[width=0.5\textwidth]{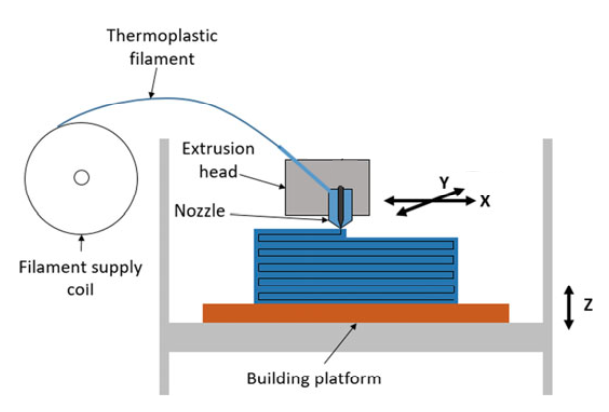}}
        \caption{Illustration of the FDM process \cite{HazratAli2021}}
        \label{fig:fdm}
    \end{figure}
    
The precise movement of both the tool head and the filament is typically controlled by stepper motors. There are primarily three common kinematic configurations in FDM printers:

\begin{itemize}
    \item Bed slinger: the print bed moves along the Y-axis, while the tool head moves in X and Z.
    \item CoreXY: the print bed moves along the Z-axis, and the tool head moves in X and Y.
    \item Delta: the tool head is suspended from three arms that move vertically to position the nozzle.
\end{itemize}

FDM is widely used due to its affordability, ease of use, and compatibility with a broad range of materials. However, depending on the material, careful temperature management is critical. Improper temperature control can lead to issues like warping or poor layer adhesion. For this reason, most printers feature temperature regulation not only for the nozzle but also for the print bed and in some cases even for the entire build chamber.

Overall, FDM is a highly versatile process, supporting a wide variety of materials and use cases. This flexibility has led to the development of numerous printer designs optimized for different needs and applications.

\section{Common issues in the printing-process} \label{common issues}
The aim of this work is to predict printing issues of a given 3D model. Thus possible issues, which are directly influenced by the geometry of the part have to be assessed. So what do we consider to be a printability issue? The idea is to compare the finished printed part with the input file. Any noticeable differences, which are not inherent to the process like for example layer-lines, are considered an issue. Issues commonly found in literature are warping, bed-adhesion, overhangs, warping, structural robustness and resolving thin walls \cite{Dave2021, Keshavamurthy2021, DBLP:journals/corr/abs-2010-12930}.

\begin{figure}[ht!]
        \centerline{\includegraphics[width=0.5\textwidth]{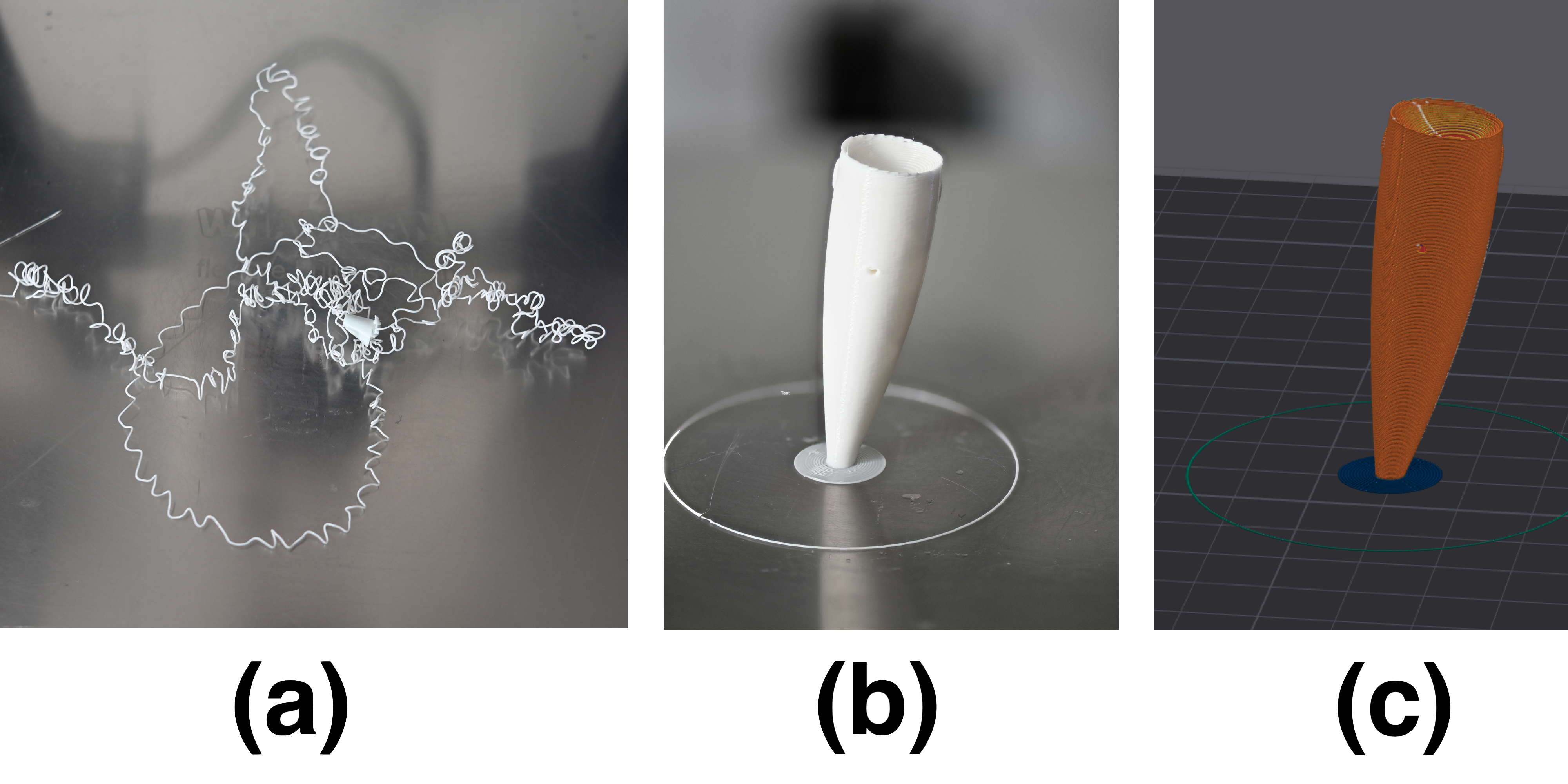}}
        \caption{Problematic bed-adhesion, \textbf{a} failure, \textbf{b} successful print of the same part with brim, \textbf{c} preview of the print with brim in blue}
        \label{fig:Adhesion}
    \end{figure}

\textbf{Bed-adhesion:}
Every print starts with a layer printed directly on the print bed. Getting this right is crucial for preventing the part from tipping over or moving during printing. A part tipping over usually goes unnoticed by the printer so it will continue to deposit material in the air. This usually ends up filling the print bed with a ball of thin strings also referenced as spaghetti. Bad adhesion can have many causes, like a dirty print bed, unsuitable material, a bad calibration of the distance from the nozzle to the bed. But interesting for this work is the problem of having a part with a small bottom surface compared to its height. If the user is aware of this issue a brim as can be seen in Fig. \ref{fig:Adhesion}c, or changing the parts orientation can help for example.  

\textbf{Warping:}
As described before warping is inherent to building a part layer by layer. As the material cools down and contracts at different points in time. For many parts this is still not a problem, as the effect usually is only noticeable on large flat parts. To reduce the effect the user can choose a suitable material, increase the bed temperature, if possible increase the temperature of the print chamber and take measures to increase the bed-adhesion.   

\begin{figure}[ht!]
        \centerline{\includegraphics[width=0.5\textwidth]{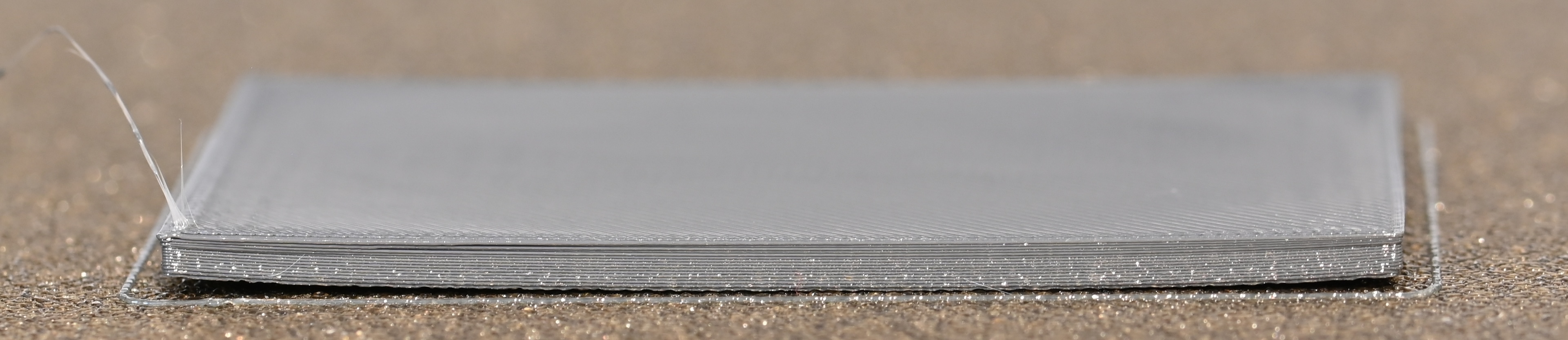}}
        \caption{Warped printed part}
        \label{fig:warp}
    \end{figure}

\textbf{Overhangs:}
Overhangs are sections of a part where the next layer extends beyond the already printed structure. This is usually fine up to an angle of around $45^\circ$ , when part cooling is sufficient. Steeper overhangs can cause serious issues ranging from a bad outer surface of the part to collisions with the print head. Most overhangs can be made printable by adding support structures. But this once again adds the need for postprocessing. And some geometries are also not suitable for non soluble supports. As they have to be accessible, the part has to withstand braking off the supports and tall supports have a risk of tipping over or braking during printing.  

\begin{figure}[ht!]
        \centerline{\includegraphics[width=0.5\textwidth]{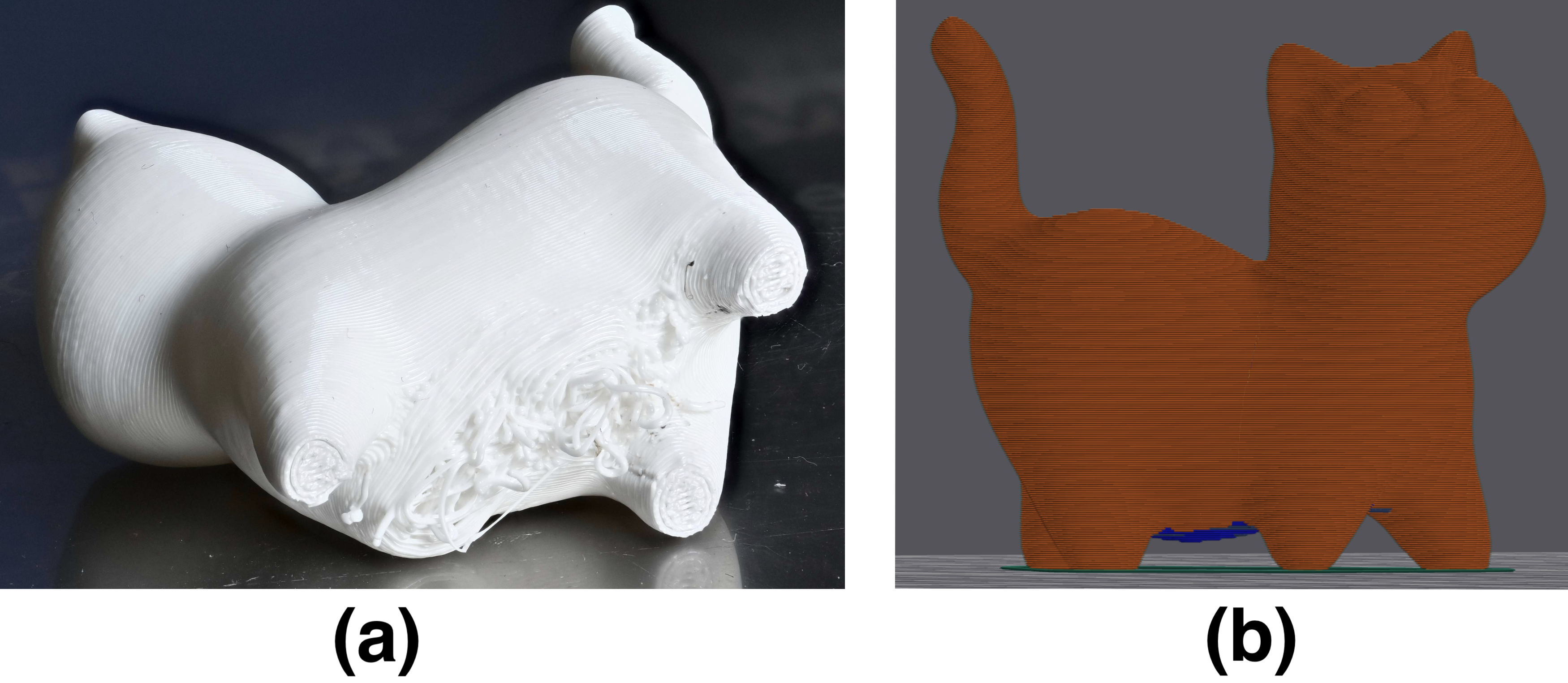}}
        \caption{Problematic overhang, \textbf{a} bottom view of the printed part, \textbf{b} preview of the same part with the problematic section highlighted in blue}
        \label{fig:overhang}
    \end{figure}

\textbf{Fine structures:}
The resolution a printer can print at is in large parts determined by the diameter of the nozzle. Typically this is $0.4$mm. For any geometry smaller than that there are primarily two options. Any geometry smaller than the line width is ignored and not printed or the geometry is printed but at increased thickness. Both approaches can lead to undesirable effects, so  it can be important that the user decides which is better for the specific part. Detecting this issue requires a dense pointcloud additionally, it is rare to find models with this issue. So the decision was made to not include it this research.   

\begin{figure}[ht!]
        \centerline{\includegraphics[width=0.5\textwidth]{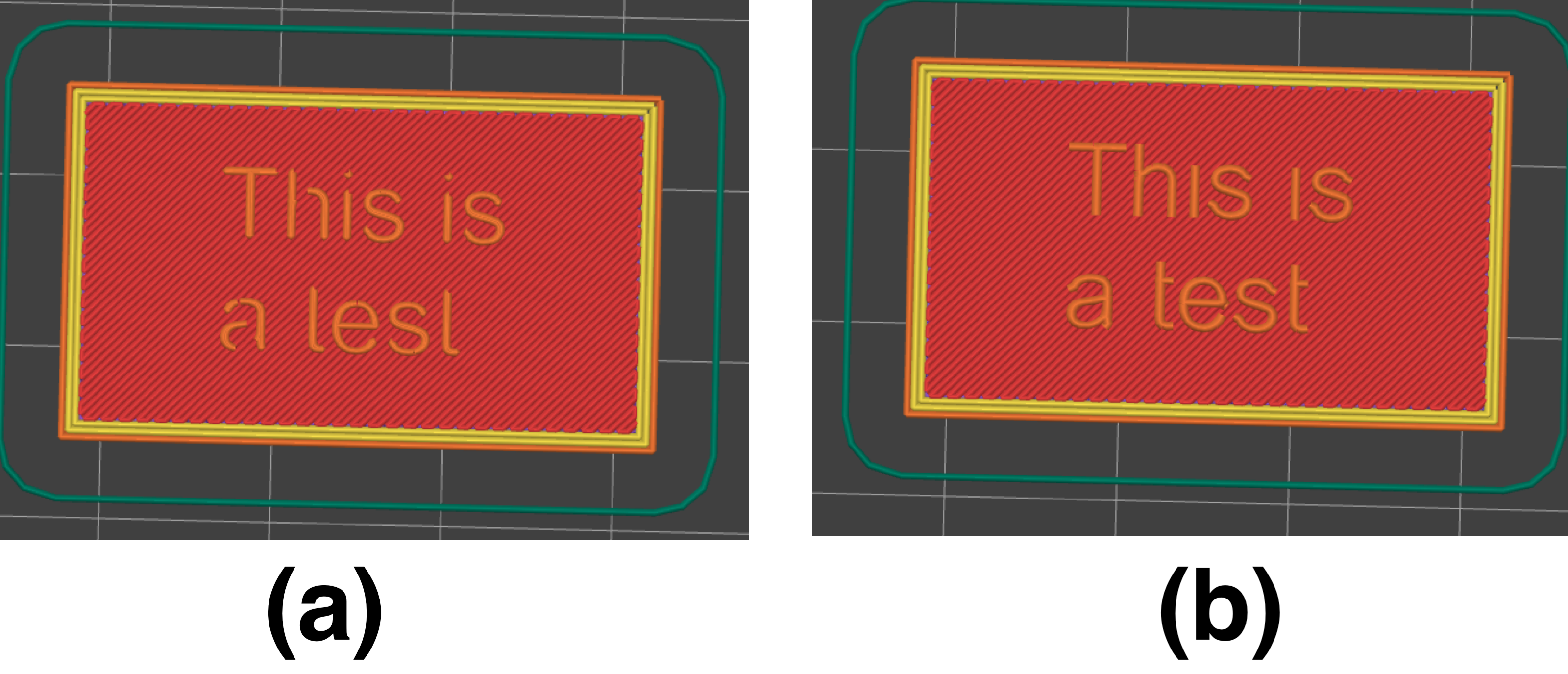}}
        \caption{Printing fine text, \textbf{a} default, \textbf{b} with setting \textit{detect thin walls}}
        \label{fig:finestruct}
    \end{figure}

\textbf{Fragile section:}
Layers are inherent to this 3D printing process and come not only with an aesthetic drawback but also form a mechanical weak point. This means tall parts with a small cross section turn out significantly weaker than flat and wide parts.

\section{Models} \label{Models}
In this section, we explore the two Machine Learning based approaches we used to assess these common issues with 3D models intended for FDM 3D printing.  One model is based on classification, while the other outputs a continuous score to reflect varying degrees of printability and is therefore of the Regression type. The following subsections describe the data collection, preprocessing pipeline, model architecture, and training process for both variants.

    \subsection{Data collection} \label{Datacollection}
    As noted the concept of evaluating 3D models based on their printability has not been formally addressed in prior machine learning  works yet, therefore no suitable labeled dataset existed for our purposes. Consequently, we curated a custom dataset tailored to the specific challenges of FDM 3D printing.
    
    The success of a machine learning model largely depends on the quality of the dataset it is built on. Therefore, dataset construction should be carried out with careful consideration of its impact on the model's performance. Poor-quality datasets can lead to inaccurate, unreliable, or biased results \cite{datasetquality}. 
    
     To ensure meaningful outcomes during training, we constructed a high-quality, balanced dataset that reflects a representative distribution of common printability issues, while including only models originally intended for FDM 3D printing. For this purpose, we sourced models from \href{https://www.printables.com/}{Printables.com}, one of the largest repositories of 3D-printing related models. Using the platform's random filter, we sampled a diverse and unbiased subset of models. To ensure consistency and avoid complications during preprocessing, several constraints were applied during dataset curation. First, only single part models were included to eliminate issues related to floating geometry, which could interfere with point cloud sampling (see Section~\ref{Preprocessing}). Similarly, models containing excessively small features were excluded, as these are often underrepresented or lost during surface sampling. Finally, it was assumed that all models were provided in their optimal orientation for FDM 3D printing. This typically involves aligning the largest flat surface with the build plate to maximize bed adhesion while minimizing overhangs.
     Then each model was manually labeled according to our predefined printability criteria.
     
    This stage marks the first point of differences between the classification and regression model. Our goal was to evaluate four of the most common printability issues in FDM 3D printing, as outlined in Section~\ref{common issues}: severe overhangs, inadequate bed adhesion, warping, and structural fragility. 
    
    For the classification model, each collected STL file was manually assigned to one or more of these four issue categories. An additional category labeled \textit{fine} was introduced for models that exhibited none of the listed problems. In cases where a model exhibited multiple issues, it was included in all corresponding categories to maintain class balance during training.
    In contrast, the dataset for the regression model was constructed by scoring each model individually on all four criteria. Each issue was rated on a continuous scale from 0 to 1, where 0 represents ideal printability and 1 indicates a completely unprintable condition with respect to that specific issue.\\
    In total, we collected around 150 unique 3D models and split them before labeling for each model. This was done while trying to be as consistent as possible and reducing bias. Despite our efforts to standardize the labeling process, it is important to note that many printability issues such as fragility or bed adhesion problems can be inherently subjective, depending on user experience, printer setup, and material choice. A detailed discussion of this subjectivity and its implications is provided in Section~\ref{Validation}.

    \subsection{Preprocessing} \label{Preprocessing}
    After dataset curation all the data should be preprocessed to further improve the models results.
    Our models operate on point cloud data, requiring each input STL file to be transformed from a mesh-based representation into a set of 3D points. 
    
    A point cloud is a collection of points in 3D space, typically defined by their Cartesian coordinates $(x, y, z)$. Unlike mesh or voxel representations, point clouds provide a lightweight and flexible format that captures the surface geometry of a model while being invariant to mesh complexity and resolution. In case of our printability evaluation, this representation is particularly advantageous: it simplifies preprocessing and reduces memory usage. Adding to this the already mentioned issues with 3D Printing are mostly only surface related problems, meaning the efficiency of point clouds not needing to represent the full volume compared to for example voxel based methods is even further advocating for a point cloud based approach.
    
    Furthermore one is not only able to push the number of points causing an increase in resolution without needing more memory but even utilizing less sampled points compared to voxels the end result will have a better representation of small and fine features. This is essential for the task at hand in terms of evaluating the fragility of a 3D printed part.
    
\begin{figure}[ht!]
        \centerline{\includegraphics[width=0.5\textwidth]{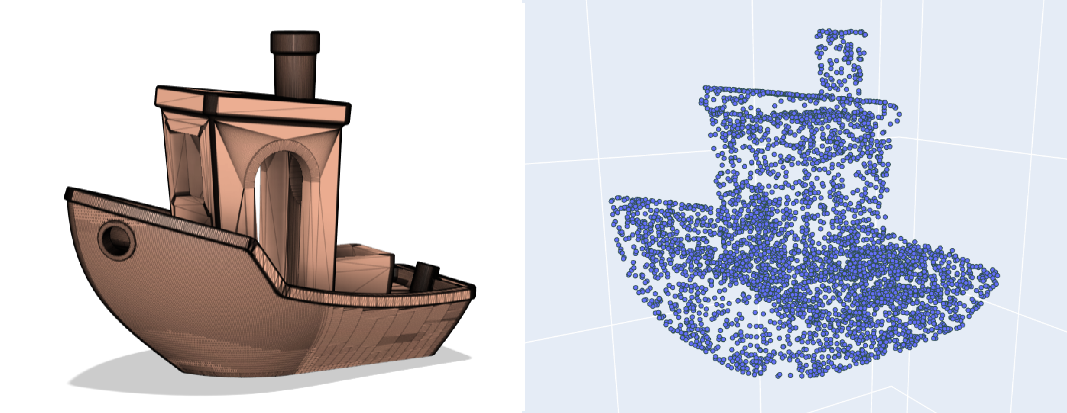}}
        \caption{Comparison of STL file with point cloud.}
        \label{fig:stl-pointcloud comparison}
    \end{figure}

    To convert STL files into point clouds, we sample points weighted by the area of the triangles across the surface of the model (see Figure~\ref{fig:stl-pointcloud comparison}). Ensuring a uniform distribution of points over the surface. Surface areas with greater polygon density receive proportionally more points, ensuring that larger features are adequately represented. This area-weighted sampling helps preserve the geometric fidelity of the model while avoiding over representation of small triangles that might otherwise bias the network. This works to some extent but as seen in Figure~\ref{fig:stl-pointcloud comparison} it does not ensure a true uniform distribution of all the points over the surface of each model.
    However, this step also introduces challenges: models composed of multiple disconnected parts that are close together may fuse into one singular part diminishing the value for training. As mentioned in \ref{Datacollection} such models where therefore excluded from the curated dataset. Further limitations of this approach include extremely small features. These can lead to uneven sampling, loss of these features which also may lead to artifacts in the resulting point cloud. To mitigate these issues, models with such characteristics were also excluded during dataset construction (see Section~\ref{Datacollection}). 
    
    Once sampled, each point cloud undergoes additional preprocessing to ensure invariance and consistency. First, the point cloud is centered by subtracting the centroid of all points, ensuring translation invariance while preserving relative distances and scale. Unlike some 3D learning pipelines, we intentionally avoid normalizing the point cloud to unit scale, as absolute size and proportion can be informative for printability (e.g. small overhangs vs. large structural elements). To achieve rotational invariance around the build plate’s vertical axis, each model is randomly rotated around the Z-axis during training. This step reflects the fact that, in practice, many 3D prints can be freely rotated on the build plate without affecting print success. After these transformations, the final input to the model is an unordered set of $N$ sampled points, represented as a matrix of shape $N \times 3$, where each row corresponds to a points $(x, y, z)$ coordinates.

    \subsection{Architecture} \label{Architecture}
    For our work, we required a model architecture that directly processes point cloud data while remaining flexible and easy to adapt to our specific task. We therefore adopted the PointNet architecture \cite{charles_pointnet_2017} due to its robustness, efficiency, and modular design.
    
    \begin{figure}[ht!]
    \centerline{\includegraphics[width=0.5\textwidth]{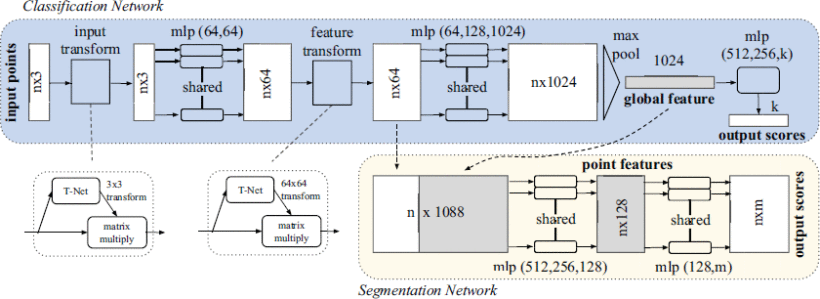}}
    \caption{PointNet Architecture \cite{charles_pointnet_2017}. For our porpuses only the classification part was used.}
    \label{fig:PointNet_architecture}
    \end{figure}
    
    PointNet operates on a set of $n$ unordered points, each represented by $(x, y, z)$ coordinates. To ensure permutation invariance, meaning the same output regardless of the order of input points, the architecture applies a shared multi-layer perceptron (MLP) to each point independently. These MLPs first map each point to a 64-dimensional space and later to 1024 dimensions. This transformation is illustrated very well in Fig.~\ref{fig:PointNet_architecture}.
    
    The resulting high-dimensional point features are then aggregated using a max pooling operation to produce a single global feature vector, which captures the overall shape of the input point cloud. 
    To maintain geometric consistency, PointNet includes a spatial transformer network (T-Net) before each MLP. These T-Nets learn affine transformations to normalize the input space. However, for our application, this step is less critical since the 3D models in our dataset are already properly aligned.
    
    The global feature vector is then passed through a final MLP to produce the model output.
    
    Originally, PointNet was designed for 3D object classification tasks using the ModelNet10 and ModelNet40 datasets \cite{ModelNet}. We adapted the architecture for evaluating the 3D printability of models targeted at fdm printers. As described in Section~\ref{Datacollection}, we pursued two approaches:
    \begin{enumerate}
        \item \textbf{Regression-based model:} To score the presence of specific printability issues, we modified the final MLP to output four continuous values—one for each type of issue (see Section~\ref{common issues}). We removed the softmax layer to preserve these as raw, interpretable scores.
    
        \item \textbf{Classification-based model:} Here, each model was labeled based on the most significant issue it exhibited. A challenge arises when a model has no issues; since classification does not naturally express absence of a problem, we introduced an additional class called \textit{fine}, as detailed in Section~\ref{Datacollection}. This resulted in a total of five output neurons. As with the regression model, the softmax layer was omitted to yield raw scores, which can still be interpreted as a ranking across categories.
    \end{enumerate}
    
    These modifications maintain the core structure and advantages of the original PointNet while enabling it to tackle the unique demands of 3D printability evaluation.

    \subsection{Training} \label{Training}   
    Both the classification and regression models were trained using a dataset of 96 samples, with an 80/20 split for training and validation, respectively. A batch size of 12 was used throughout training to balance memory efficiency and gradient stability. We employed the Adam optimizer with a fixed learning rate of 0.00025, which provided stable convergence across experiments. To prevent overfitting, a dropout rate of 0.3 was applied to the final fully connected layers. This prevents the model of relying on small portions of the neurons and ensures more generalization. 
    For the classification model, the loss function consisted of the Negative Log-Likelihood (NLL) loss, augmented by a regularization term that encourages the learned feature transformation matrix in the T-Net module to remain close to orthogonal. This regularization term helps preserve geometric invariance and improve generalization. For our classification model the Loss function is therefore the same as the one in the original paper \cite{charles_pointnet_2017}. The regression model shares the same training configuration, but replaces the NLL loss with a mean squared error (MSE) criterion, allowing the model to learn continuous-valued printability scores across all four defined outputs.
    
\section{Validation} \label{Validation}
With the validation we try to answer two questions. How accurate are the models on the training data and the additional validation data labeled by us? The other question is how much the results are biased by our personal experience.
\begin{enumerate}
    \item \textbf{Classification-based model:} 
    In order to validate the accuracy of the model with respect to the dataset the accuracy was plotted both for the training data and the validation data over the number of epochs trained (Fig. \ref{fig:accuracy}). The accuracy is determined by checking if the category with the highest value matches the label. The accuracy is the amount of correct guesses divided by models in the dataset.  
    
    \begin{figure}[ht!]
        \centerline{\includegraphics[width=0.5\textwidth]{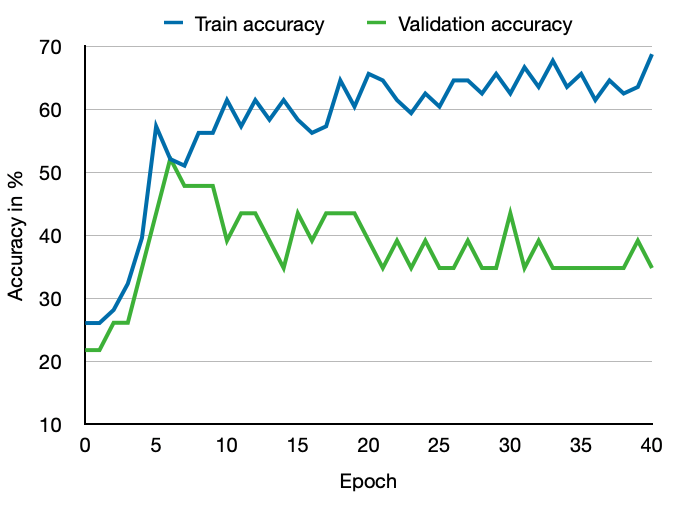}}
        \caption{Accuracy of the classification model on training and validation data}
        \label{fig:accuracy}
    \end{figure}
    
    Fig. \ref{fig:accuracy} shows that after about 10 epochs clear overfitting occurs. Even with overfitting the train accuracy does not exceed 70\%. This is likely due to the fact that some of the models in the training data were placed in multiple categories. This means it is impossible for the accuracy to be higher than 50\% for those models. Overall this suggest to train the model for 10 epochs. Taking a look at the output ratings manually it does seem, that the results get better up about 15 epochs. Once again it shows, that the very small amount of validation data may not show the full story.
    
    \item \textbf{Regression-based Model:}  
    To assess the performance of the regression-based model, we monitored the training and validation cost across 40 training epochs. The results are visualized in Fig.~\ref{fig:regression-cost}, which plots the cost values over time.
    
    \begin{figure}[ht!]
        \centerline{\includegraphics[width=0.5\textwidth]{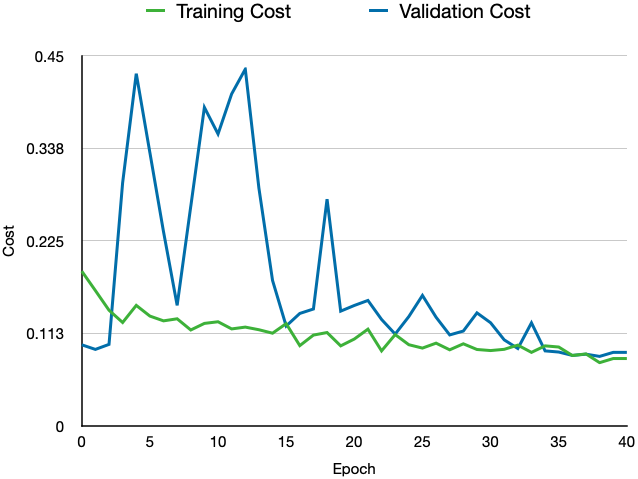}}
        \caption{Training and validation cost curves of the regression model over 40 epochs. Values above 0.5 were treated as outliers and linearly interpolated for better visualization.}
        \label{fig:regression-cost}
    \end{figure}
    
    The figure clearly illustrates a relatively smooth and consistently decreasing training cost, suggesting that the model effectively minimized the loss on the training set. In contrast, the validation cost exhibits substantial fluctuations particularly during the early and middle stages of training. These sharp spikes are indicative of instability in generalization, which can be attributed to the small validation set (20 samples) used during training. With such a limited evaluation set, individual outlier samples can have a disproportionately large impact on the computed validation loss, especially in a regression context where prediction errors are continuous rather than categorical.
    
    To improve visual evaluation, validation cost values exceeding 0.5 were considered statistical outliers and replaced with interpolated values based on their neighboring points. This smoothing helps reveal the general trend of the model's learning process without being dominated by extreme fluctuations.
    
    Despite the noise in the validation signal, an overall downward trend in both curves is noticeable across the 40 epochs. This indicates that the model is learning meaningful representations of the data. However, extending training beyond 40 epochs did not lead to further improvement in validation performance; instead, it resulted in rising validation cost, signaling the onset of overfitting. Based on this observation, the final model was trained for 40 epochs, representing a trade-off between learning capacity and generalization performance.
    
    These results underscore the importance of larger and more balanced datasets when training regression-based models, which tend to be more sensitive to outliers and label noise than classification models. Furthermore, the validation behavior highlights the necessity of careful tuning of hyperparameters and regularization when operating in data-constrained environments.      
\end{enumerate}
  
    \subsection{User survey}
    For this survey, three users experienced in FDM 3D printing were asked to evaluate five models by identifying potential printability issues and ranking them by significance. This approach allows for a direct comparison between the users’ rankings and those generated by the AI model. Additionally, it provides insight into the variance between human evaluators. The results are summarized in Fig.\ref{fig:survey} in the appendix.

    Overall, the user responses were largely consistent. For instance, the most significant issue identified matched across all three users for 3 out of the 5 models. However, only one model received an identical ranking from all participants, indicating that not every model has a single, objectively correct ranking of issues. This variability is likely due to the fact that the perceived severity of printability problems does not depend solely on the model’s geometry, but also on external factors such as the material used, printer type, print settings, and the intended use of the part.

    Despite this subjectivity, the results suggest that issue rankings can be generalized to some extent, though discrepancies between different evaluators including AI models are to be expected.

    When comparing the two AI models, we find that their outputs are quite similar, despite being trained on largely different datasets. In general, both AI models align reasonably well with human evaluations. However, there are notable exceptions. For example, the fourth model was consistently rated as unproblematic by all human participants, yet both AI models flagged it as having multiple issues. Similarly, for the first model, both AIs identified overhangs as problematic, whereas none of the users mentioned this.

\section{Limitations} \label{Limitations}
Our results show that adapting a model originally designed for object classification can give promising results when applied to evaluating printability issues in 3D models. However, this approach also presents several notable challenges.

One major limitation is the difficulty of labeling training data. A single part may have multiple printability issues, yet with the classification model, each training sample can only be labeled with one issue. This prevents the loss function from making full use of the model’s ranked output. This constraint likely negatively impacts training effectiveness. While increasing the size of the dataset could help mitigate this problem, doing so is non-trivial, as labeling is a manual and time-consuming process. 

Another challenge lies in the inherent data requirements of machine learning models. AI models typically require large volumes of data to reliably identify subtle yet critical features. For example, bed-adhesion depends heavily on having a flat bottom surface. If this surface is even slightly angled, such that parts of it are more than one layer height (typically 0.2mm) away from the build plate—it effectively becomes an overhang, increasing the risk of print failure.

\begin{figure}[ht!]
    \centerline{\includegraphics[width=0.5\textwidth]{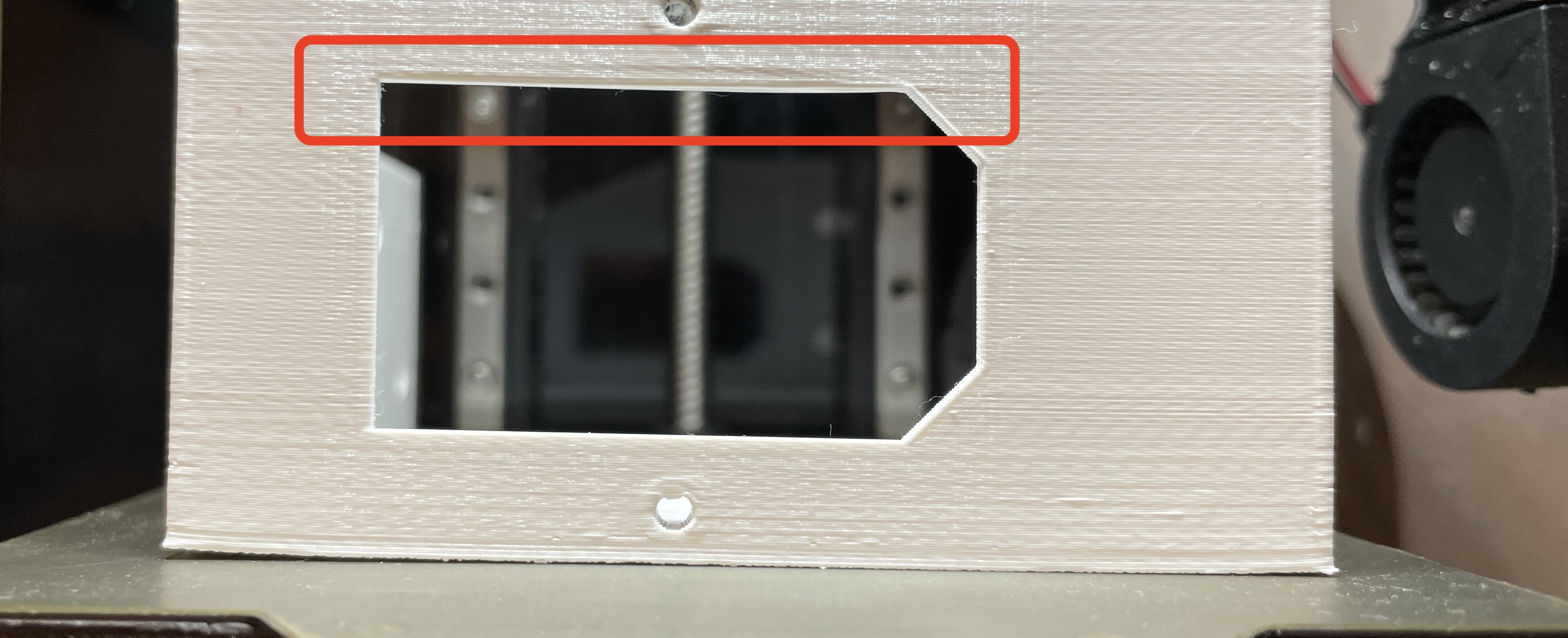}}
    \caption{Part with a bridged section (marked in red)}
    \label{fig:bridge}
    \end{figure}

Even with a significantly larger and more refined dataset, full generalization remains unlikely. The severity of a printability issue often depends on context that may not be available from geometry alone. For instance, certain design techniques, such as bridging, can allow features that would normally be problematic (like horizontal spans with no support underneath) to print successfully. Bridging works as long as the start and end points of each extruded line are anchored to previously printed material, as illustrated in Fig.\ref{fig:bridge}. On the other hand additional variables like material, settings or use case of the part can also play a role.
 
\section{Outlook}
Overall, the classification model proved to be functional and capable of identifying common printability issues, but it remains far from optimal. The primary limiting factor is the small dataset size, compounded by the inherent subjectivity of the manual labeling process. This subjectivity, while addressed in part through repeated labeling and validation (see Section~\ref{Validation}), still introduces inconsistencies that likely constrain model performance. Future work should focus on significantly expanding the dataset and establishing more objective, rule-based labeling guidelines to reduce ambiguity and increase the reliability of both training and evaluation. This would, in turn, enhance the classification model's accuracy and robustness. 

The regression model, while based on the same core architecture and training pipeline, poses additional challenges. Regression tasks generally require larger and more diverse datasets to learn meaningful continuous mappings, especially when targeting complex and very subjective criteria like the common issues in FDM printability. Unlike classification, where categorical boundaries can provide some robustness to noise, regression models are more sensitive to inconsistent labeling and limited coverage. Therefore, any improvement to this model will likely require not only more data, but also more precise scoring protocols. Using semi automated, deterministic scoring algorithms might help for some common issues like bed adhesion but could be difficult to implement for more complex issues like overhangs (see section~\ref{Limitations}).

Both models would likely benefit from transfer learning via fine-tuning the original PointNet model \cite{charles_pointnet_2017}, pretrained on the ModelNet40 dataset \cite{ModelNet}. Transfer learning could improve generalization without the need for further large-scale data collection. This could involve replacing only the final prediction head of the original PointNet while keeping earlier layers fixed or retraining the model for a small amount of epochs, both leading to faster and more efficient convergence. In future iterations, more advanced architectures such as PointGST~\cite{Pointgst} or other transformer-based models could be employed to further improve performance, especially for capturing complex 3D geometries of each model. PointGST proposes a promising way to freeze and fine tune only small parts of a large and advanced point cloud based model by introducing a lightweight, trainable Point Cloud Spectral Adapter to fine tune parameters in the spectral domain. This will eventually make it a promising foundation for models trained on limited datasets, such as the one presented in this work.

Moreover, integrating segmentation based approaches such as the segmentation branch of PointNet (see Fig.~\ref{Architecture}) offers an exciting avenue for extension. This would enable localized printability analysis, where problematic regions (e.g., fragile overhangs or poor bed-contact areas) could be visualized directly on the model. Such feedback would not only aid in training explainable models but also enhance usability by guiding users in making targeted design improvements.

From a practical standpoint, user accessibility remains a core motivation of this work. Integrating our evaluation model directly into slicing software would make it easier for non experts to identify potential issues before printing, thereby reducing failed prints and material waste. Coupled with segmentation based visual feedback, this would create a powerful tool for ease of use in 3D printing and lowering the barrier to entry for beginners.

A further enhancement could involve the use of rotational sampling, where each model is evaluated in multiple orientations to identify the most optimal print orientation. By scoring different orientations on their expected printability, the system could recommend the best orientation automatically, this would be an especially valuable feature for inexperienced users or for complex models where intuition may be misleading.

In conclusion, while our current models provide a promising foundation for automated printability evaluation, there is substantial room for improvement through larger datasets, better labeling protocols, transfer learning, and more advanced architectures. Future work should focus not only on improving predictive performance but also on maximizing usability and integration with existing 3D printing slicers. By continuing in this direction, we hope to make reliable printability assessment a standard and accessible part of the 3D modeling and printing pipeline.


\bibliographystyle{ieeetr}

\onecolumn
\appendix

\section{Appendix}
\begin{figure}[h]
    \centering
    \includegraphics[width=0.7\textwidth]{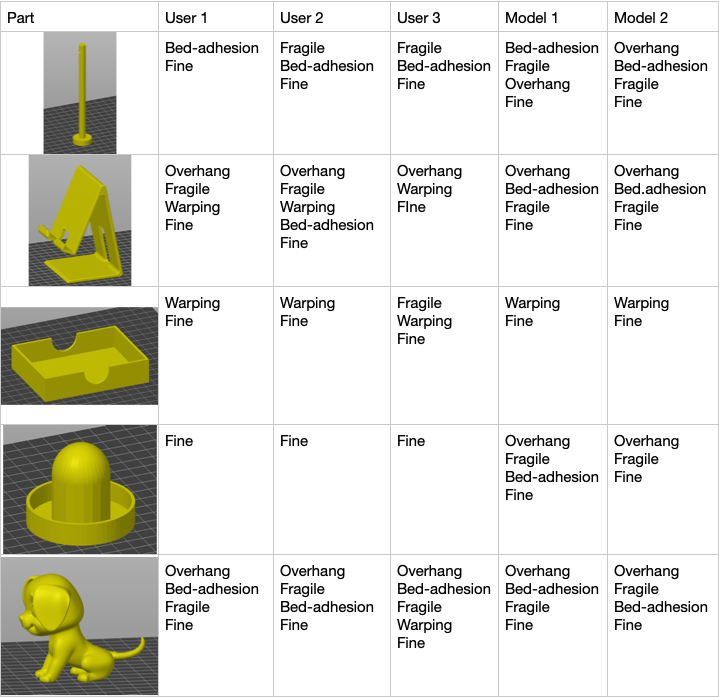}
    \caption{Results of the user survey, Model 1 is the AI trained on single-label data, Model 2 is the trained with multiple labels per part}
    \label{fig:survey}
    \end{figure} 

\end{document}